# Creativity in translation: machine translation as a constraint for literary texts

Ana Guerberof-Arenas* and Antonio Toral**

*University of Surrey/University of Groningen
**University of Groningen

*Abstract*

This article presents the results of a study involving the translation of a short story by Kurt Vonnegut from English to Catalan and Dutch using three modalities: machine-translation (MT), post-editing (PE) and translation without aid (HT). Our aim is to explore creativity, understood to involve novelty and acceptability, from a quantitative perspective. The results show that HT has the highest creativity score, followed by PE, and lastly, MT, and this is unanimous from all reviewers. A neural MT system trained on literary data does not currently have the necessary capabilities for a creative translation; it renders literal solutions to translation problems. More importantly, using MT to post-edit raw output constrains the creativity of translators, resulting in a poorer translation often not fit for publication, according to experts.

*Keywords*

creativity, machine translation post-editing, literary translation, creative shift, acceptability

*Introduction*

As part of a larger project,[1] we look at the effect of neural machine translation (NMT), on translators' creativity. We previously analysed creativity of translations of a short fictional text[2] that allowed us to also test narrative engagement (Busselle and Bilandzic 2009), enjoyment (Hakemulder 2004), and translation reception from a cohort of Catalan readers. The results showed that human translation (HT), judged by one professional literary translator, scores higher for creativity than MT post-editing (PE) and, unsurprisingly, higher than MT (Guerberof-Arenas and Toral 2020).

---

[1] CREAMT: https://cordis.europa.eu/project/id/890697

[2] The modified version of Murder in the mall by Nuland (1995) tells the story of an escaped psychiatric patient who, in trying to kill a nine-year-old girl, stabs a homeless person to death who has instinctively intervened to protect the child; the event is witnessed and retold in part by the girl's mother.

Our previous experiment, however, had certain limitations. First and foremost, the text itself, in which action and emotion prevailed over style. Second, we tested our research questions only in one translation direction (English-to-Catalan). Third, the translations were judged by only one professional translator.

To account for these limitations, we have slightly modified the methodology and present here the results of a subsequent experiment. Firstly, we summarize previous work related to creativity and translation before outlining our adjusted methodology and presenting the results from our analysis of the texts and of the interviews with nine professional literary translators and drawing our conclusions.

*Related Work*

Creativity is a complex process, so much so that, although often mentioned in relation to artificial intelligence as an intrinsic human quality, it is seldom defined or analysed. The psychology branch that specializes in creativity is surprisingly new. Experts tend to locate the starting point at Guilford's keynote speech at the 1950 annual meeting of the American Psychological Association (Guilford 1950). Since then the discipline has flourished (Runco 2014), and there is now a consensus that creativity "requires both originality and effectiveness" (Runco and Jaeger 2012, 93) or "something novel and appropriate, from a person, a group or a society" (Sawyer 2006, 33). This implies that for something to be labelled as creative, it needs to be new (even if it involves the reshaping of an existing object or idea), but it also has to be of value for the context in which it is created (there is an element of practicality, although this is not always a prerequisite). Creativity is "the drive to come up with something that is new and surprising and that has value" (Sautoy 2019, 4). It is not only a mental process, it is a cultural and social phenomenon that is universal and "may be expressed not only in the visual arts, music or poetry, but also in everyday activities" (Rudowicz 2003, 276) such as mathematics, machine learning, business and innovation, and, related to this work, translation (Malmkjær 2019).

In Translation Studies, creativity has also been an elusive concept (O'Sullivan 2013)

often linked to the dichotomy between literal and non-literal translation, and hence to the general concept of translation shifts (translations that depart from the literal). Rojo explains that "creative solutions have to be novel and depart from conventional translation behaviour, but should also render meaning accurately and give solutions appropriate to a certain textual, situational and cultural context" (Rojo 2017, 353). Bayer-Hohenwarter and Kussmaul (2020, 312) explain that "a creative translation is a translation that often involves changes (as a result of shifts) when compared to the source text, thereby bringing something that is new and also appropriate to the task that was set, i.e. to the translation assignment (to purpose)". Kussmaul (1995), based on psychology studies, applies a four-phase model to the creative process: A) preparation related to the comprehension of the ST, B) incubation related to the search for ideas in the known world, C) illumination related to the production of these ideas, and D) evaluation related to the revision of the ideas. These phases do not necessarily occur in sequential order. For this research, we are particularly interested in the operationalization of the creativity concept used by Bayer-Hohenwarter (2009; 2010; 2011; 2012; 2013). She defines creativity as involving a combination of four dimensions: Acceptability (the translation meets the requirements of the brief), Flexibility (the use of *creative shifts* as opposed to a *reproduction*, a literal rendering of the source text), Novelty (how unique a translation is in comparison to others) and Fluency (the number of translation solutions provided for one problem by one translator). She then develops a scoring system to rate translations with a view to understanding competence and training in translation.

Some of the most recent research in NMT and literary translation (earlier research is covered in Guerberof-Arenas and Toral 2020) investigates three aspects: A) the usability of NMT in the translation of literary texts, B) its effect on translators and translation students, and C) professionals' opinions on technology in different countries.

In A, two studies look at Google Translate (GT) in the translation of an Agatha Christie's novel from English to Dutch (Fonteyne, Tezcan, and Macken 2020; Tezcan, Daems, and Macken 2019). They use two methods, error categorization by means of an error taxonomy and computational linguistics analysis: lexical richness, local cohesion, syntactic and stylistic differences. The encouraging results show that 44% of the sentences do not contain errors. As a follow up to this study, Webster et al. (2020) look at the translation of four classic novels from English to Dutch using GT and DeepL. Their results are less optimistic than previous results with only 23 % (GT) and 21% (DeepL) of sentences found to be without errors. In our view, these results point to the fact that, when translating literature, the genre and author's style plays a fundamental role, since these texts are not as homogeneous as technical texts.

In B, four studies find a priming and constraining effect when post-editing MT. Sahin and Gürses (2019) in English to Turkish, Kenny and Winters (2020) in English to German, Mohar, Orthaber and Onic (2020) in English to Slovene, and Kolb (2021) in English to German all find that the PE versions are either less creative or diminish the translators' voice, and that MT has a priming effect on translators. Not all studies find that the PE version contains more errors, but the types of errors identified are different than those find in the HT versions (often linked to style).

In C, three studies (Daems 2021; Ruffo 2021; Slessor 2020) analyse the use of technology among literary translators using surveys in the UK, Canada, the Netherlands and Belgium. The results show that literary translators are not reluctant to use technology such as word-processing, online dictionaries, terminology databases, or search engines, but are less inclined to use CAT tools or MT solutions, mainly because they feel they are not suitable for literary translation, pinpointing to the fact that creativity and lexical and syntactical variation are important in this domain. They also find that some reluctance might be caused by not being familiar with recent technological developments, not enough training, and perhaps due to age.

*Research Methodology*

The CREAMT project is articulated in two main axes. The first one aims to identify creativity in different translation modalities. The second axis seeks to identify readers' narrative engagement and enjoyment, and to gather data on translation reception. Here we present the results of the first axis, driven by the research questions:

**RQ1** Can we quantify the creativity in texts translated by humans as well as in those produced with the aid of machines?

**RQ1.1** Is the creativity level different according to translation modality?

In this first axis we are interested in looking at the literary translators' product when working with different modalities to answer our primary question about creativity. However, and given that the number of participants in this part of our project is limited (4 translators and 5 reviewers), we analysed the data available from different angles to give our readers a more informed view of how creativity in translation is affected by MT, not only in the final product, the translated text, but also in the translation process. The different data points complement the information and help to answer our questions and explain the findings in a richer and more holistic way.

*Source Text*

The quest for a text that met our experiment requirements was arduous. The text had to meet the following conditions: A) higher creative potential than the one used in our previous experiment (Guerberof-Arenas and Toral 2020), B) not present in our engine's training data to not artificially favour MT, C) engaging enough to measure reading experience, D) short enough to be read in under an hour and E) not too outdated so that all generations could engage with it. We selected an intriguing story by Kurt Vonnegut:[3] 2BR02B. It was first published in 1962 in the magazine *Worlds of If Science Fiction* and was later included in the collection *Bagombo Snuff Box* (Vonnegut 1999). This science fiction story is set in a futuristic world where death has been eradicated and people live forever, the only caveat is that for one person that is born, another one must die. Vonnegut centres the story on several characters, a father to be, a doctor, a government official, a hospital orderly and a painter. These characters exemplify different attitudes towards the world they are immersed in.

The text contains 123 paragraphs, 234 segments and 2,548 words. To our knowledge, this story has not been translated into Catalan nor Dutch. The text has a Flesch Reading Ease index of 79 and a Flesch-Kincaid Grade Level of 4.6. These indexes are designed to indicate how difficult a text is to understand; in this case, the scores indicate that it is fairly easy to read for an English speaker.

*Machine Translation*

---

[3] Project Gutenberg Literary Archive Foundation https://www.gutenberg.org/files/21279/21279-h/21279-h.htm

The MT modality was based on the output of state-of-the-art literary-adapted neural MT systems based on the Transformer architecture (Vaswani et al. 2017). The system for English-to-Catalan was trained with more than 130 novels in English and their translations into the target language, as well as around 1000 books in the target language, and over 4 million sentence pairs from a variety of domains collected by Softcatalà[4] (Toral, Oliver, and Ribas-Bellestín 2020). The system for English-to-Dutch was trained with parallel novels, namely 500 novels in English and their translations in Dutch, which amount to around 5 million sentence pairs (Toral, van Cranenburgh, and Nutters 2021). The training data did not contain the text used for the experiment nor any text from Kurt Vonnegut.

*Translation Process*

The text was translated from English to Dutch (NL) and Catalan (CA). These languages were selected for two reasons: convenience of sample and availability of a literary-adapted MT engine already trained and tested. The HT and PE versions were provided by four professional translators who specialize in literary translation.[5] The CA translators are identified herein as T1 and T2, and the NL translators as T3 and T4.

As in our previous experiment, the translators used the PET post-editing tool (Aziz, Castilho, and Specia 2012), segmented by paragraph. To reduce the risk that a reader would engage more with one translator's work because they prefer that translator's style, this method was followed: for example, T1 translates the first half and post-edits the second half while T2 post-edits the first half and translates the second half. The text is then aggregated per modality. When creating the HT and PE texts, certain inconsistencies were found. These were not caused by the translators, but by the fact that the text came from two different sources. The decisions to choose one term over another were based on frequency or order of appearance in the translated text.[6]

---

[4] https://github.com/Softcatala/en-ca-corpus

[5] To recruit the translators, two websites which offer a database of literary translators were used (Expertisecentrum Literair Vertalen and Associació d'escriptors en llengua catalana). Some translators recommended others based on availability. The translators that took part in our previous experiment were contacted as reviewers for CA. The NL translators came from The Netherlands.

[6] Appendix A can be found at http://github.com/AnaGuerberof/CREAMT

*Translators*

The four professional translators, two females and two males, had between 5 and 15+ years of experience and had translated between 15 and 18+ novels during this time. A post-assignment questionnaire was used to gather the translators' opinions on the quality of the MT output and their perceived effort. Overall the CA translators rated the NMT engine better[7] (from Slightly good to Moderately good) than the NL translators (between Moderately bad to Neither good nor bad), and T4 was particularly unhappy with the quality of the engine (Moderately bad overall). T3 and T4 had stated in the pre-task questionnaire that they did not have PE experience and T4 explained that he did not like using MT because the proposals were poor. Finally, the translators clearly prefer translating to PE. In terms of effort, the opinions are diverse among translators: while the translators perceived they were faster in PE, their cognitive and technical effort was also higher in this modality.

*Review process*

The five reviewers[8] were also professional literary translators, four females and one male, they had between 5+ and 15+ years of experience and had translated between 15 and 40 novels, individually or as part of a team. The CA reviewers are identified herein as R1, R2 and R3, and the NL reviewers as R4 and R5. Only one of them reported having experience in PE (R4).

The reviewers were given three texts and instructions to fill in the forms (see Measuring Creativity). Each text had a code (A, B and C) but they were blind to how the texts had been translated. Text A was the PE text; Text B, the MT text; and Text C, the HT text. After the assignment a short questionnaire was used to gather the reviewers' overall opinions on the quality of the translations (see the Results section).

*Measuring Creativity*

---

[7] The translators were asked to rate the output in a 7-point Likert-type scale that ranged from Extremely good to Extremely bad. All results can be found at:
https://github.com/AnaGuerberof/CREAMT/tree/main/questionnairesaxis1

[8] Three reviewers for CA and two reviewers for NL: one reviewer came from the Netherlands, and one from Belgium but resided in the Netherlands. Three reviewers were engaged also for NL, but one of them did not complete the assignment following the instructions, and the data had to be discarded.

As explained in Related work and as defined in our previous project (Guerberof-Arenas and Toral 2020), creativity combines acceptability (i.e. something of value, fit for purpose) and novelty (i.e. new, original). Creativity cannot be only defined, nor quantified, only in terms of novelty. If this were the case, the limitations a literary translator finds when creating something new that fits the meaning of the ST and the target culture would not be considered, that is, translators could come up with new and attractive solutions disregarding the intention of the ST as a matter of course. In the same way, MT could be rewarded when coming up with "hallucinations"[9] (Lee et al. 2018) and could be classified as creative. Therefore, acceptability was operationalized by the number and severity of errors in the translated texts, and the praises (Kudos), and novelty (or flexibility), by the number of creative solutions (shifts) provided for a given problem posed by the ST.

*Acceptability*. To rate the translated texts, the reviewers used the harmonized DQF-MQM Framework[10]. The errors are classified according to the following categories: Accuracy, Fluency, Terminology, Style, Design, Locale Convention, Verity and Other. Reviewers also had to annotate the severity of each error: Neutral (0 points for repeated errors or preferences), Minor (1 point), Major (5 points) and Critical (15 points). Kudos was used for exceptionally good translation solutions.[11] The reviewers were sent instructions on how to perform each task and they were instructed to contact the researchers if something was unclear.[12] The total number of points, not errors, was considered for this experiment.

*Units of creative potential (UCP).* To assess novelty, the ST was marked with UCPs. Two experienced researchers and translators annotated the ST with units that are expected to require translators to use problem-solving skills, as opposed to those that are regarded as routine units (Bayer-Hohenwarter 2011; Fontanet 2005). The annotators were given a set of instructions and a UCP classification list as follows, although they were free to mark others: A) metaphors and original images, B) comparisons; C) idiomatic phrases, D) wordplay and puns, E) onomatopoeias, F) colloquial language (cursing, slang, for example), G) phrasal verbs, H) cultural and historical references, I) neologisms, J) lexical variety (number of adjectives before the noun or use of adverbs, for example) K) expressions specific to linguistic variant (for example, American English or British English) L) unusual punctuation, M) rhyme and metrics, N) proper names, and O) treatment (formal, informal).

---

[9] Hallucinations are 'highly pathological translations that are completely untethered from the source material' (Lee et al. 2018, 1).

[10] https://www.taus.net/qt21-project

[11] https://www.taus.net/qt21-project#harmonized-error-typology

[12] The process was smooth and clear for all participants. However, the data for one NL reviewer had to be discarded because it was incomplete and could not be repaired as the task was already over budget.

The annotators identified the same 77 units (41.62 %), and 108 different units. Since translation problems could be different depending on the language and the annotator criteria, it was decided to aggregate them, thus there were 185 UCPs in total (out of 123 paragraphs).

*Novelty* The reviewers were instructed to classify the UCPs in the target texts (TTs) in a spreadsheet especially created for the project. For this classification the previous version (based on Bayer-Hohenwarter 2011) was adjusted with a subclassification of the units (based on Pedersen 2011)[13] as shown below.

—Reproduction: All translations that reproduce the UCP with the same idea or image, even if they are acceptable. They can be then classified into Retention, Specification, Direct Translation or Official Translation.

—Omissions: When a term or expression from the UCP is omitted in the TT. An omission can be subclassified as creative or a shortcut solution.[14]

—Errors: If the translation is not acceptable (contains too many errors), then it can be marked as Not Applicable (NA).

—Creative shifts (CS): All translations that deviate from the ST in any of the following ways:

   —"Abstraction" refers to instances when translators use more vague, general or abstract TT solutions. An abstraction could be subclassified into Superordinate Term or Paraphrase.

   —"Concretisation" refers to instances when the TT evokes a more explicit, more detailed and more precise idea or image. A Concretisation could be classified into

---

[13] Data can be found at https://github.com/AnaGuerberof/CREAMT/tree/main/creativeshifts

[14] A creative omission is the one that might be required in the target culture.

Addition or Completion.

—"Modification" refers to instances when translators use a different solution in the TT (e.g. express a different metaphor without the image becoming more abstract or concrete). A Modification could be subclassified into Cultural, Situational or Historical.

*Creativity Scoring*. Acceptability and novelty are combined into a single score. The result is multiplied by 100 to express it as a percentage.

$$creativity\ score = (\frac{\#CSs}{\#UCPs} - \frac{\#error\ points - \#Kudos}{\#words\ in\ ST}) * 100$$

*Interviews*

A post-task semi-structured interview was carried out to gather more information about the translation and reviewing activities. The participants were each interviewed by video call[15] for under one hour. The questions were designed to elicit responses from participants, e.g. "Could you give me your general views on translating/reviewing this text?", "Where would you say that you/the translator had to be more creative?", "What did you think of the MT proposals?", "What did you think of Text A, Text B, Text C?".

## Results

This section is divided in three parts: A) results from the translation process, B) results from the review, and C) the analysis of interviews.

*Results on translators' changes and effort*

This section reports on the translation and PE effort and on the differences between the translated and post-edited texts.

*HTER*

---

[15] This research was conducted at the height of the COVID-19 pandemic so no alternative was available.

The final translations were analysed using HTER, Human-Targeted Translation Error Rate (Snover et al. 2006), an automatic score that reflects the number of edits performed on the MT output normalized by the number of words in the sentence. The closer HTER is to 0, the fewer changes performed. Figure 1 shows the HTER results in boxplots according to the modality (PE and HT) and the languages involved (CA and NL) using NMT as the hypothesis.[16]

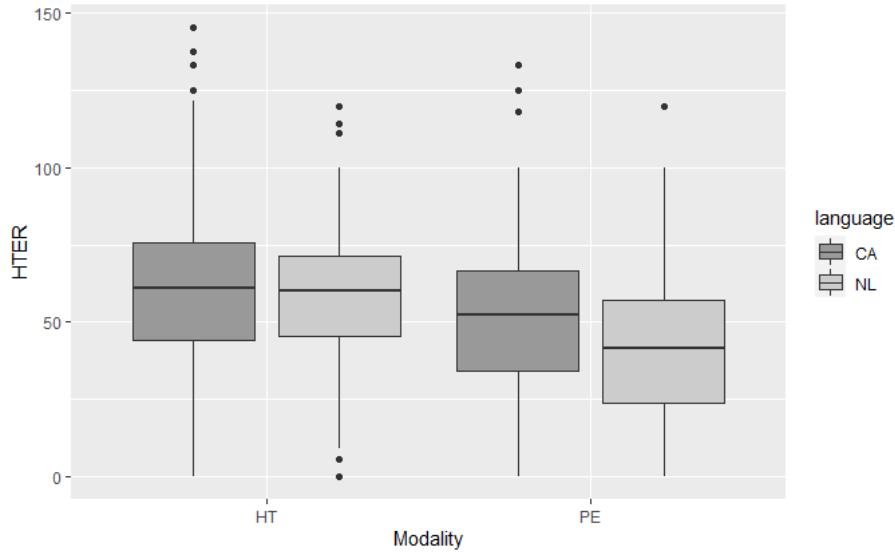

Figure 1: HTER per language and modality

The figure shows that, as expected, the HT is further from the MT output in both languages (CA M = 61.96; NL M = 58.81) than the PE (CA M= 54.72, NL M = 41.82), but also that the NL translation is always closer to the MT output, especially for the PE modality. To see this in more detail, Table 1 shows the HTER per modality (PE and HT) and translator.

| Translator | PE | HT | Language |
|---|---|---|---|
| T1 | 52.43 | 59.04 | CA |
| T2 | 56.97 | 64.93 | CA |
| T3 | 45.45 | 64.86 | NL |
| T4 | 38.25 | 52.66 | NL |

Table 1: HTER results for PE and HT per translator

---

[16] The reference (NMT) is the translated text (paragraph) that the hypothesis (HT or PE) is compared against.

For PE, CA translators made more changes than their NL counterparts, with T4's PE being the closest to the MT proposal. This is interesting since both NL translators (T3 and T4) rated the MT output the lowest, yet still they made fewer changes during PE. A possible explanation is that, since they were new to the PE task, they might have felt that they had to preserve more of the NMT output resulting in an increased level of frustration and lower ratings of the output. The data also indicates that the quality of the output is not suitable for production (Schmidtke and Groves 2019) and that even using a customized and improved NMT engine for CA, the HTER results are higher for the PE modality than in our previous study (Guerberof-Arenas and Toral 2020). This could be related to the fact that this new ST was more creative, and therefore more changes were required for an optimal final quality.

*Translation effort according to PET*

When analysing the data from PET, nine NL segments showed unusually high edit times. Those paragraphs were checked with the translators and we concluded that some were left open even when not worked on, causing several extreme outliers. Since these segments skewed the data, they were removed from the dataset for all four translators. Each segment represents a paragraph, so we removed nine out of 123 paragraphs (19, 82, 85, 91, 93, 94, 96, 105 and 107). Table 2 shows a summary of the data, N is therefore 114 segments.

| Modality | Language | Total edit time (s) | Total keystrokes | Total pauses > 1s | Total annotations[17] |
|---|---|---|---|---|---|
| HT | CA | 22392 | 25005 | 1548 | 360 |
| HT | NL | 26938 | 19298 | 1498 | 276 |
| PE | CA | 18879 | 12294 | 1028 | 359 |
| PE | NL | 31234 | 8449 | 1279 | 273 |

Table 2: Translation effort according to PET

Table 2 shows some differences between modalities but also between languages. The average edit time between modalities (HT M = 216.36, PE M = 219.80) is very close. However, NL took longer on average to translate (CA HT M = 196.42, NL HT M =

---
[17] Here "annotations" represent how many times the translators opened and closed segments in PET.

236.22) and to post-edit than CA (CA PE M =165.81, NL PE = 273.98).

The NL translators took longer in both activities, especially PE, although they implemented fewer edits in the output (as seen in the HTER section). This could point to the quality of the MT output, but also to translators' experience in PE and the natural speed of each individual translator. Since the sample is small, general conclusions cannot be drawn, but it helps to understand our dataset and the subsequent results on creativity.

The average number of keystrokes between modalities (HT M = 194.03, PE M = 90.98) is very different. Both languages have a higher number of keystrokes when translating (CA HT M = 219.34, NL HT M = 169.28) than for PE (CA PE =107.84, NL PE = 74.11), and CA has a higher average overall. It is logical that translators made more keystrokes when translating from scratch since there is no preexisting typed text. However, as we have seen, less typing did not mean less time for NL. Although previous research in PE of literary texts shows a reduction in both variables (Toral, Wieling, and Way 2018), other research in creative text shows that a reduction in keystrokes does not necessarily mean a reduction in time or effort (Koponen et al. 2020). The explanation for this can be manifold: the language combination, the type of text, the MT engine, and the number and type of participants.

The average number of pauses (Lacruz, Denkowski, and Lavie 2014) is higher in HT than in PE (HT M = 13.36, PE M = 10.12). Both languages have a higher number of pauses when translating (CA HT M = 13.58, NL HT M = 13.14) than for PE (CA PE M = 9.02, NL PE M= 11.22). This data seems to indicate that translating was more cognitively demanding for translators than PE. In PE research, a lower cognitive effort is perceived as an indicator that MT has helped translators to reduce their effort. However, we hypothesize that, in our case, a higher cognitive effort indicates that translators 'think harder' for a more creative solution because pauses are an indication of the 'incubation' period in creative translation (Kussmaul 1991).

If we look at number of annotations (i.e. the times that translators enter a paragraph, where one annotation means that it was opened once) we observe that EN-CA translators went back to the segments more times, which could be an indication of a higher level of revision. It is interesting to note that long revisions might also be an indicator of the creative process (Heiden 2005; Rojo 2017). However, the time spent on the segments might not be a 100% reliable representation of the real time spent to elaborate the final translation in the translators' mind (incubation) outside the tool.

If we compare these results with the translators' perceived effort, as per post-task questionnaire, the technical (keystrokes) and cognitive effort (pauses) are different. Perhaps the effort of doing an enjoyable activity (HT) is perceived as lower than that in a less enjoyable activity (PE) even if the results show the opposite.

These results on effort are in line with recent work on creativity from English to Chinese (Vieira et al. 2020).

*Results from the review*

This section covers the results from the review to assess creativity. Even if the reviewers were unaware of the modalities, we include them in the results shown for clarity.

*Post-task questionnaire.* The reviewers were asked to rate the modalities. The reviewers were unanimous in raking HT as an Extremely good translation, MT as an Extremely bad translation, and PE as a Neither good nor bad translation. In the PE case, the CA translation received a higher rating, between Slightly good to Moderately good, than the NL translation, Slightly bad to Moderately bad.[18] It is interesting to note that the NL PE was the closest to the MT proposal (see Table 1), even if translators took longer to process this modality, possibly indicating that the closer a translation is to MT, the lower the quality of the translation.

*Acceptability*. The 5 reviewers annotated the three texts (HT, PE and MT).[19] Table 3 summarizes the results by error points, i.e. considering the severity of the errors, not only the number of errors.

| Error category | HT | | PE | | MT | |
| --- | --- | --- | --- | --- | --- | --- |
| | CA | NL | CA | NL | CA | NL |
| Accuracy | **12** | **43** | **25** | **106** | **238** | **337** |
| Design | 0 | 0 | 1 | 0 | 1 | 0 |
| Fluency | **11** | **16** | **25** | **30** | **73** | **47** |
| Locale convention | 0 | 1 | 1 | 1 | 2 | 1 |
| Other | 0 | 1 | 0 | 5 | 7 | 11 |

---

[18] https://github.com/AnaGuerberof/CREAMT/tree/main/questionnairesaxis1

[19] https://github.com/AnaGuerberof/CREAMT/tree/main/acceptability

| Style | 7 | 13 | 14 | 71 | 89 | 127 |
|---|---|---|---|---|---|---|
| Terminology | 0 | 2 | 0 | 4 | 6 | 1 |
| Verity | 0 | 0 | 1 | 5 | 0 | 5 |
| Total | **31** | **75** | **66** | **221** | **416** | **528** |

Table 3: Average error point according to modality and language

In line with our previous findings (Guerberof-Arenas and Toral 2020), MT contains more errors than the other two modalities put together. However, in this case, PE also contains a strikingly higher number of errors than HT (double the errors in CA and three times in NL), even if the same translators worked on the HT and PE. The reason behind this could be the type of text chosen (the current ST has a high number of UCPs). However, the number of participants is low, so the ability of each participant to post-edit or to translate is also a factor.

The categories with the most errors are Accuracy, Fluency and Style. In this case, NL shows higher numbers than CA in terms of Accuracy in all modalities. This could be related to the reviewers, perhaps the NL reviewers were stricter, or the NL translators made more errors. This is also in line with what translators mentioned in the interviews, i.e. they find the PE task limits their creativity, and is technically and cognitively more demanding.

Table 4 shows the average kudos that the reviewers gave to the translations. HT scores higher in both languages, and NL is highly praised in the HT modality.

|  | HT | | PE | | MT | |
|---|---|---|---|---|---|---|
|  | CA | NL | CA | NL | CA | NL |
| Average kudos | 5 | 31 | 3 | 8 | 0 | 1[20] |
| Total | **36** | | **11** | | **1** | |

Table 4: Average kudos according to modality and language

Inter-annotator agreement (Fleiss Kappa) was run in the different modalities and languages according to the errors marked. Although reviewers agreed on the overall quality of the translations, they do not always agree if there was an error or on the type of error in a given sentence.[21] Tables 5 and 6 show a small sample of errors found by all the reviewers for comparison.

---

[20] "'Hooray,' said Wehling emptily" was translated by the engine as "'Hoera,' zei Wehling wezenloos". Wezenloos means vacant or expressionless.

| Source Text | Translation | M | R1 | R2 | R3 |
| --- | --- | --- | --- | --- | --- |
| **The floor** was paved with **spattered dropcloths.** | **Aquí i allà el terra** estava cobert amb **teles protectores.** | HT | **El terra** estava **tot** cobert amb teles protectores.<br><br>Accuracy/Mistranslation/Minor | Aquí i allà el terra estava cobert amb teles protectores **totes esquitxades de pintura**.<br><br>Accuracy/Omission/Minor | Aquí i allà el terra estava cobert amb teles protectores **esquitxades**.<br><br>Accuracy/Undertranslation/Minor |
| "What **man in my shoes wouldn't be happy?**" said Wehling. | —Qui **estaria** content al meu lloc? —va preguntar Afflick | PE | —Qui **no** estaria content al meu lloc? —va preguntar Afflick<br><br>Accuracy/Mistranslation/Major | —Qui **no** estaria content al meu lloc? —va preguntar Afflick**,**<br><br>Fluency/Punctuation/Minor | —Qui **no** estaria content al meu lloc? —va preguntar **l'**Afflick<br><br>Fluency/Grammar/Minor & Style/Unidiomatic/Minor |
|  | —Quin home de les meves sabates no estaria content? | MT | **—Qui no estaria content, en el meu lloc?** —va dir en Wehling.<br><br>Accuracy/Mistranslation/Major | **—Que estaria content algun home en la meua situació?** —va dir en Wehling.<br><br>Accuracy/Unstranlated/Major | —Quin home [de] **no estaria content al meu lloc?** —va dir en Wehling.<br><br>Accuracy/Mistranslation/Major |

---

[21] Values below 0.40 represent poor agreement and values above represent fair to good agreement. For CA HT, kappa = 0.14, CA PE kappa = 0.22 and CA MT kappa = 0.34. For NL HT, kappa = 0.24, NL PE kappa = 0.28 and NL MT kappa = 0.50. Neutral errors are included in this calculation.

| | —va dir en Wehling. | | | |

Table 5: Sample of errors and reviewers in CA

| Source Text | Translation | Modality | R4 | R5 |
|---|---|---|---|---|
| His camouflage was perfect, since the waiting room had a **disorderly** and demoralized air, too. | Zijn camouflage kon niet beter, aangezien er in de wachtkamer ook een sfeer van slordigheid en ontmoediging hing. | HT | Zijn camouflage kon niet beter, aangezien er in de wachtkamer ook een sfeer van **wanorde** en ontmoediging hing. Style/Unidiomatic/Minor | Zijn camouflage kon niet beter, aangezien er in de wachtkamer ook een sfeer van **rommeligheid** en ontmoediging hing. Accuracy/Mistranslation/Minor |
| **Back in the days when** people **aged** visibly, **his age would have been guessed at thirty-five or so.** | In de tijd toen mensen nog zichtbaar oud werden, zouden ze gedacht hebben dat hij een jaar of vijfendertig was. | PE | **In de tijd dat** mensen nog zichtbaar **verouderden**, zouden ze gedacht hebben dat hij een jaar of vijfendertig was. Fluency/Grammar/Minor & Accuracy/Mistranslation/Minor | In de tijd toen mensen nog zichtbaar **ouder** werden, zouden ze **hem op een jaar of vijfendertig hebben geschat.** Accuracy/Mistranslation/Minor (2) |
| "**Good gravy**, no!" she said. | "Mooie jus, nee!" zei ze. | MT | **'Goeie grutten**, nee!' zei ze. Accuracy/Mistranslation/Major & Fluency/Punctuation/Neutral | **'Goeie grutten**, nee!' zei ze. Accuracy/Mistranslation/Major |

|  |  |  |  |  |
|---|---|---|---|---|

Table 6: Sample of errors and reviewers in NL

*Creative Shifts*. Table 7 shows the average number of reproductions and creative shifts in the 185 UCPs, as classified by the reviewers.

| Classification | HT | | PE | | MT | |
|---|---|---|---|---|---|---|
| | CA | NL | CA | NL | CA | NL |
| Abstraction | 6 | 28 | 6 | 25 | 2 | 9 |
| Concretisation | 15 | 9 | 11 | 6 | 1 | 1 |
| Modification | 61 | 42 | 49 | 32 | 11 | 17 |
| **Total CS** | **82** | **79** | **65** | **63** | **15** | **26** |
| NA | 1 | 1 | 2 | 6 | 23 | 19 |
| Omission | 4 | 2 | 8 | 1 | 9 | 2 |
| **Reproduction** | **98** | **104** | **110** | **116** | **139** | **139** |

Table 7: Reproduction and CSs in the three modalities

The results show that HT has the highest number of CSs followed by PE and lastly by MT. If the type of shifts is observed, HT is clearly above PE when it comes to Modification, where translators changed the ST to adapt it to the target culture. This type of CS was more prominent in this experiment than in our previous experiment, because the type of text required to draw a parallel dystopia into the CA and NL cultures. Inter-annotator agreement (Fleiss Kappa) was run in the different modalities and languages to see if reviewers agree on those UCPs where there was a CS. The agreement was fair to good in all cases except for NL MT, for which it was poor.[22] Tables 8 and 9 show the CS classification for the strings corresponding to Tables 6 and 7.

| Source Text | UCP | Modality | Classification | Subclassification | Reviewer |
|---|---|---|---|---|---|
| The floor was paved with spattered dropcloths. | spattered dropcloths | HT | Omission | Shortcut | R1 |
| | | HT | Omission | Shortcut | R2 |
| | | HT | Omission | Creative | R3 |

---

[22] Values below 0.40 represent poor agreement and values above represent fair to good agreement. For CA HT, kappa = 0.46, CA PE kappa = 0.43 and CA MT kappa = 0.53. For NL HT, kappa = 0.42, NL PE kappa = 0.50 and NL MT kappa = 0.37. If the UCP was marked as NA by one of the reviewers, the UCP was eliminated from the calculation.

| "What man in my shoes wouldn't be happy?" said Wehling. | What man in my shoes | PE | Modification | Situational | R1 |
|---|---|---|---|---|---|
| | | PE | Modification | Situational | R2 |
| | | PE | Modification | Situational | R3 |
| "What man in my shoes wouldn't be happy?" said Wehling. | What man in my shoes | MT | NA | NA | R1 |
| | | MT | Reproduction | Retention | R2 |
| | | MT | Reproduction | Direct Translation | R3 |

Table 8: Sample of CS in CA

| Source Text | UCP | Modality | Classification | Subclassification | Reviewer |
|---|---|---|---|---|---|
| His camouflage was perfect, since the waiting room had a disorderly and demoralized air, too. | disorderly and demoralized air | HT | Abstraction | Paraphrase | R4 |
| | | HT | Reproduction | Direct Translation | R5 |
| Back in the days when people aged visibly, his age would have been guessed at thirty-five or so. | Back in the days | PE | Reproduction | Direct Translation | R4 |
| | | PE | Abstraction | Superordinate Term | R5 |
| "Good gravy, no!" she said. | Good gravy | MT | Reproduction | Direct Translation | R4 |
| | | MT | NA | Not applicable | R5 |

Table 9: Sample of CS in NL

*Creativity Score*. Table 10 shows the creativity scores for each modality and language according to the formula presented in the Methodology section.

| | CA | | | NL | | |
|---|---|---|---|---|---|---|
| TM | HT | PE | MT | HT | PE | MT |
| # CS | 82 | 65 | 15 | 79 | 63 | 26 |
| Units | 185 | | | | | |
| # Errors Points | 31 | 66 | 416 | 75 | 221 | 528 |
| # Kudos | 5 | 3 | 0 | 31 | 8 | 1 |
| ST words | 2548 | | | | | |
| Creativity Score | 43 | 33 | -8 | 41 | 25 | -7 |

Table 10: Creativity scores according to modality

HT shows the highest creativity score in both languages: HT is 10 points higher than PE and 51 points higher than MT in CA, and HT is 16 points higher than PE and 48 points higher than MT in NL. These results confirm our previous findings where HT creativity was higher than the other modalities. MT is not only less creative than a translation done or post-edited by translators, since it has not only more errors but also fewer creative shifts, but using MT as part of the translation (modality PE), also results in a less creative literary translation.

*Interviews with translators and reviewers*

The transcriptions from the nine interviews were analysed using Nvivo 12 Pro.[23] Firstly, from the semi-structured interviews, categories were drawn within two separate superordinate activities: Translation (6 categories and 28 subcategories) and Review (7 categories and 14 subcategories). Further, emerging themes were observed, and nodes were merged from both translators and reviewers into a single thematic classification of 6 themes and 19 subthemes. Finally, Figure 2 shows the final eight-theme structure followed by a discussion on these themes.

| Final Themes | Files | References |
|---|---|---|
| MT can only be partially useful in literary translation | 9 | 209 |
| Creativity is a problem solving selection process | 9 | 205 |
| The proposals, as in post-editing, acts as a constraint to creativity | 9 | 189 |
| The delicate equilibrium of reviewing | 5 | 189 |
| The ideal tool is human centered | 9 | 151 |
| The technology or modality impact depends on the type of reader | 9 | 134 |
| The final translation as a product of many collaborative steps | 6 | 82 |
| The grim future of translation and technology | 4 | 26 |

Figure 2: Themes from interviews

---

[23] https://github.com/AnaGuerberof/CREAMT/tree/main/interviewsaxis1

1. *MT can only be partially useful in literary translation*

The participants mentioned that MT could be useful, but perhaps in those parts of the text that are mechanical and straightforward. It does not yet have an "eye for context" and "no eye for style" and it misses the coherence of a whole text, only partially working at a sentence level. Because it is "too literal", there are many "huge errors", "mistranslations", causing MT to become impractical. A translator would look at the whole text and if they fall short in one part, they will compensate in another part, but "would a machine ever try and compensate?".

On occasions, however, they were surprised at the quality of the MT: "some of the sentences didn't really look like if they had been just machine translated and never touched", "it came up with quite a nice proposal and maybe I wouldn't have thought of this myself", "it was a lot better than I had expected". One of the translators said that she uses MT to generate ideas in moderation as to not "intoxicate your brain".

They mentioned that an MT version could be useful for "branches of the editorial world that don't demand the creativity that literature does", or even passages within a novel. Some thought that it would be better than having no version at all, for example, in Catalan, but others saw many dangers in misrepresenting the authors with a poor MT translation and they were unsure if the authors' intended message would get across.

2. *Creativity is a problem-solving process*

The participants define creativity as solving problems with solutions that are unexpected, "surprising" or "brilliant". During this process, it is apparent that at times translators must think harder, the effort is higher, "he or she has not settled for the first thing that has crossed his or her mind" to provide "another formulation", that "goes beyond the English" and that creates a "natural", "real", "beautiful", "fluent" style in the target text. At the same time, the solution must fit into the "context" and style given by the author with the objective of giving "the reader the experience […] of the original text". Creativity in translation is not necessarily related to the author's own creativity in the ST, but to "little sentences and natural phrasing" in the target text that bring the ST closer to the target culture, which requires the translator to "imagine the world" the author is depicting, to "understand the rules and how that universe works".

They also mention that creativity is related to "thinking outside the box" or "lateral thinking", "associative thinking", "to fill in a gap", but also to "intuition", "a gut feeling" because translation is a balancing act, the translator acts as a mediator between author and reader and between cultures. This "in-betweenness" space is defined thoroughly in Ruffo's research (2021).

Finally, creativity seems to be that part in literary translation that translators have more fun doing: "that's the part of what makes it such a great job as literary translator […] if the machine can do it, then I am just the editor", "If we take that away from our intellectual capabilities it will only make us dumber and I don't see the point". This is in line with research that shows that creativity is one of the most positive feelings humans have (Sawyer 2006).

3. *The proposals act as a constraint to creativity*

The translators thought that during PE they were "primed or conditioned" by the text they were given: "You have like a 'corset'", "it really limits the creativity", "it's not good for the translators' brain". Even if they changed the proposals, it was difficult to "step out", "because it's so easy to be fooled by a translation that is already there". It will never be like translating from scratch where translators look at the rhythm and musicality of the text "…it lacks a bit of creativity to make it believable so it's more like stating facts then then turning the facts into a story". They might think "this is all right" and end up accepting proposals that are not "genuine". Further, because during PE the focus is on a smaller piece of text, the overall coherence and the style is often lost. These comments are in line with previous research (Moorkens et al. 2018).

The proposals prevent translators from setting into motion their own mechanisms, and departing from the ST, which is one of the conditions for a creative translation according to the translators, and to initiate the creative phases as described by Kussmaul (1995). They also mentioned that this happens with any text read before "…one of the translations gets stuck in your head and it makes it more difficult to be creative". The translators found PE harder than translating on their own "It's a bigger effort cognitively and technically and creatively to improve the text".

4. *The delicate equilibrium of reviewing*

The reviewers agreed in both languages that the HT text "was a high-quality translation", "you came across these creative solutions" in what "was a very good text so it was a real pleasure to review something which was so good".

They also placed the PE "between the two extremes", in that it was too close to the English and "was less creative". Even if the translators thought that the difference would not be noticeable, "I think that the reader would consider the post-edited one and one that was done completely by me from scratch as identical", the reviewers were surprised when finding out that the same people had worked in the two modalities: "I thought that they [HT and PE] had been translated by two different people […], that's really surprising really" (sic).

Finally, in the MT text "a large-scale intervention was necessary to turn it into a readable text", "some of the words were crazy", "there was so much I think I probably missed things as well.". However, they believed that at times MT had been through some editing: "there were other points where the text could be read more or less fluently as well". Although the five reviewers agreed on the texts that they liked and disliked, it was apparent in the conversation that reviewing was a difficult task in which they juggled many variables and were constantly thinking about several options: "It was really tricky to review those texts", "I'm always second guessing myself", "it's difficult to enclose the creative shifts or the errors in one category or another", "Sometimes it was like a kind of a yo-yo or table tennis kind of thing".

5. *The ideal tool for literary translation is human centred*

The participants did not think that PE could be a viable solution for literary translation, except for certain parts of the texts. All believed there must be a human in control of the tool, not just to fix errors.

They offer recommendations for a tool: "Assisted translation for terminology issues […] to avoid jumping a paragraph or jumping a line, to make sure that the numbers, the proper names are as in the original", "to have a good database of expressions natural equivalents", "improve the idiomatic translations", "something that detects when you're being maybe too literal", "warned about certain sentence structures", "knowledge of the real world as well", "I would have the source text, the target text, so with no proposals", but also the system "would learn from my correcting […] I don't want to repeat the same kind of correction 300 times".

It seems that literary translators want to use technology but in slightly different way that is being proposed at the moment, the technology in support of the *translator* rather than a *translation* proposal (these findings are similar to those in Daems 2021; Ruffo 2021; Slessor 2020)

6. *The technology or modality impact depends on the type of reader*

The participants were aware that they, as professional translators, were a different type of reader that was used to "looking behind the scenes", but they all mentioned that the reading experience would vary depending on the type of readers: "some people […] read something which is not very, very well translated and they are happy", those that are interested in the plot and those that would focus both on the plot and the style, "a minimally sharp reader would be able to enjoy that solution", "an intelligent reader would immediately see that it's full of mistakes and inconsistencies and errors and illogical things".

Both type of readers would find the MT text challenging: it "will not be a pleasant", "they would hate [it]". For the PE text, the experience is "not terrible, terrible, but [..] it has room for improvement", "many inconsistencies and the Dutch is not really Dutch it's an in-between Dutch", while for the HT text, they all thought that "they would enjoy" it, "it would read more naturally, and it would be more entertaining". But they were also aware that readers "when they read the translation now they don't even think about it. They just think oh, this is the style of the original author."

7. *The final translation as a product of many collaborative steps*

It was clear during the interviews that the translation process is complex, and it involves several steps and collaborations.

They read the ST in a slightly different way than a normal reader "with an open mind, but I always start making notes on my first reading". The review process of a literary text is different for each translator and for each piece, but it tends to be a long, arduous process involving several rounds of reviews, some bilingual and some monolingual: "my first translations [are] always quite crude […] I put like three or four different options", "I make three reviews", "I export it to an eBook format so Mobi and I read it in my Kindle aloud", "extra review from for the 20 first pages and the 20 last pages", "sometimes I prefer to work fast and come up with the first draft that I know is far from perfect […], whereas in some cases that are more creative or that have a lot of possible pitfalls I prefer to work intensely on the sentence-by-sentence level".

The process also involves several actors "it's really impossible to just translate without somebody who looks at your text and gives feedback in some way". Translators often consult native speakers ("I always ask natives for lots of things") or the writer ("I need to ask native speakers or questions that I might need to ask the writer"), but they also tend to "decide with the editor with the publisher", "normally my editor will look at the text once, and then send it back to me", they have a professional reader or they talk to colleagues, friends and partners.

8. *The grim future of translation and technology*

Four participants made comments about the future of translation and technology which showed that this is a topic that causes some worry and anguish. They mentioned that if MT improves, big publishing houses might want to lower the price even more "I mean literary translation is already ill paid everywhere enough so that in the future they might say no, you know, 30% of your work will be solved by this translation so we will take off 30% of your rate as well, so can it be helpful? Maybe from a technical point of view but for a human perspective". "I do believe that in about 30 years we won't have translators anymore", "I think it [MT] will only get better it will get better every year".

*Conclusions*

This experiment has confirmed our previous findings (Guerberof-Arenas and Toral 2020) that creativity can be quantified and that HT scores higher for creativity than PE and MT. Clearly, the two modalities that show the highest creativity scores are those where professional translators intervene (HT and PE).

When looking at literary translations, it is not enough to look at productivity or even measurements such as technical or cognitive effort (through pauses), although these gave us interesting data points to ponder. What is the use of MT if it does not save time and only keystrokes? Further, cognitive effort does not always mean an 'unhappy' effort. Pauses in our experiment seemed to be related to the fact that translators had to think hard to come up with a creative solution, and this was something that they enjoyed. Indeed, emotion has been identified as one aspect that influence creativity (Kenny 2006; Kussmaul 1995; Rojo and Ramos Caro 2016).

In this experiment, data were obtained from two language pairs, English to Catalan and Dutch, different expert reviewers and interviews again confirm the results previously obtained. The creative shift analysis reveals that the translators in HT provide a more novel/flexible translation, less constrained by the MT output, but also with fewer errors. Further, using MT has a counter effect: during PE translators are less creative (more errors and fewer creative shifts), even if the same translators work on the two modalities. This is a cautionary tale for using MT in creative texts, even when competent translators work hard to deliver their optimal translation, or at least if MT is used as output to post-edit before the translator has had the opportunity to think of a solution (we do not know what the result might be if MT is offered as an alternative after an initial translation). The results are exacerbated with more creative STs, i.e. STs with a higher number of UCPs. We would like to finish this chapter by suggesting a definition of a creative translation based on the literature (Bayer-Hohenwarter and Kussmaul 2020; Fontanet 2005; Rojo 2017) but also using the data from our research: *creative translation is the process of identifying and understanding a problem in the source text, generating several new and elegant solutions that depart from the source text and choosing the one that best fits the target text and culture to provide the reader the same experience as that of the source reader*. From the data analysed herein, it appears that using MT (by means of PE) hinders the effectiveness of the translation process, because the translator becomes the evaluator and not the creator, and therefore the mechanisms and phases of creativity are not set into motion. To see if the reader experience is indeed altered, the second part of our project will explore the reader experience of these modalities.

*Funding information*

This project has received funding from the European Union's Horizon 2020 research and innovation programme under the Marie Skłodowska-Curie grant agreement No. 890697.

*Acknowledgements*

We would like to thank the translators Carlota Gurt, Yannick Garcia Porres, Núria Molines Galarza, Josep Marco Borillo, Scheherezade Surià, Theo Schoemaker, Roos van de Wardt, Linda Broer and Leen van de Broucke, and the annotators Tia Nutters and Gerrit Bayer-Hohenwarter for their crucial contribution to this study.

*Authors' Address*

Ana Guerberof-Arenas

https://orcid.org/0000-0001-9820-7074

Computational Linguistics Group

Center for Language and Cognition

Faculty of Arts

University of Groningen

The Netherlands

a.guerberof.arenas@rug.nl

Antonio Toral

https://orcid.org/0000-0003-2357-2960

Computational Linguistics Group

Center for Language and Cognition

Faculty of Arts

University of Groningen

The Netherlands

a.toral.ruiz@rug.nl